\pdfoutput=1

\documentclass[11pt]{article}

\usepackage[preprint]{acl}

\usepackage{times}
\usepackage{latexsym}

\usepackage[T1]{fontenc}

\usepackage[utf8]{inputenc}

\usepackage{microtype}

\usepackage{inconsolata}

\usepackage{graphicx}
\usepackage{kotex}

\usepackage{booktabs}
\usepackage{array}
\usepackage{tabularx}
\usepackage{multirow}
\usepackage{adjustbox}
\usepackage{caption}
\usepackage{subcaption}
\usepackage{enumitem}
\usepackage{array}

\usepackage{mdframed}
\mdfsetup{
    linecolor=white,
    backgroundcolor=gray!20,
    innertopmargin=4pt,   %
    innerbottommargin=4pt,   %
    innerleftmargin=4pt,   %
    innerrightmargin=4pt,  %
}

\newcommand{\beliefChoiceQ}{\textsc{BeliefQ\textsubscript{[Choice]}}}

\newcommand{\answerabilityListQ}{\textsc{Answerability Q\textsubscript{[List]}}}
\newcommand{\infoaccessListQ}{\textsc{InfoAccess Q\textsubscript{[List]}}}
\newcommand{\answerabilityYNQ}{\textsc{Answerability Q\textsubscript{[Y/N]}}}
\newcommand{\infoaccessYNQ}{\textsc{InfoAccess Q\textsubscript{[Y/N]}}}

\title{Perceptions to Beliefs: Exploring Precursory Inferences\\for Theory of Mind in Large Language Models}

\author{
Chani Jung$^{\heartsuit}$ \quad
Dongkwan Kim$^{\heartsuit}$ \quad
Jiho Jin$^{\heartsuit}$ \quad
\textbf{Jiseon Kim}$^{\heartsuit}$ \\
\textbf{Yeon Seonwoo}$^{\clubsuit}$ \quad
\textbf{Yejin Choi}$^{\diamondsuit}$ \quad
\textbf{Alice Oh}$^{\heartsuit}$ \quad
\textbf{Hyunwoo Kim}$^{\spadesuit}$ \quad
\\
\\
{$^\heartsuit$KAIST} \quad
{$^\clubsuit$Amazon} \quad
{$^\diamondsuit$University of Washington} \quad
{$^\spadesuit$Allen Institute for AI} \\ 
    \small{\texttt{\{1016chani, dongkwan.kim, jinjh0123, jiseon\_kim\}@kaist.ac.kr}} 
    \\
    \small{\texttt{yseonwoo@amazon.com, yejin@cs.washington.edu, alice.oh@kaist.edu, hyunwook@allenai.org}}
}

\begin{document}
\maketitle
\begin{abstract}
While humans naturally develop theory of mind (ToM), the capability to understand other people's mental states and beliefs, state-of-the-art large language models (LLMs) underperform on simple ToM benchmarks. 
We posit that we can extend our understanding of LLMs' ToM abilities by evaluating key human ToM precursors---\textit{perception inference} and \textit{perception-to-belief inference}---in LLMs. 
We introduce two datasets, Percept-ToMi and Percept-FANToM, to evaluate these precursory inferences for ToM in LLMs by annotating characters' perceptions on ToMi and FANToM, respectively.
Our evaluation of eight state-of-the-art LLMs reveals that the models generally perform well in perception inference while exhibiting limited capability in perception-to-belief inference (e.g., lack of inhibitory control).
Based on these results, we present PercepToM, a novel ToM method leveraging LLMs' strong perception inference capability while supplementing their limited perception-to-belief inference. 
Experimental results demonstrate that PercepToM significantly enhances LLM's performance, especially in false belief scenarios.

\end{abstract}

\section{Introduction}

Humans interact with others in various social situations using~\textit{theory of mind} (ToM), the cognitive capability to understand other's mental states \citep[e.g., beliefs, desires, and thoughts;][]{premack1978does}.
While ToM is naturally developed for humans in childhood, large language models (LLMs) are known to exhibit inconsistency in ToM tasks~\citep{van-duijn-etal-2023-theory, trott2023large}. 
Despite some early reports of successful cases \citep{whang2023nytimes, street2024llms}, studies have shown that even state-of-the-art LLMs significantly lag behind human performance in ToM tasks, particularly in false belief tests~\citep{le-etal-2019-revisiting, kim-etal-2023-fantom, gandhi2023understanding, wu-etal-2023-hi, shapira-etal-2024-clever}.

\begin{figure}[t]
  \includegraphics[width=\columnwidth]{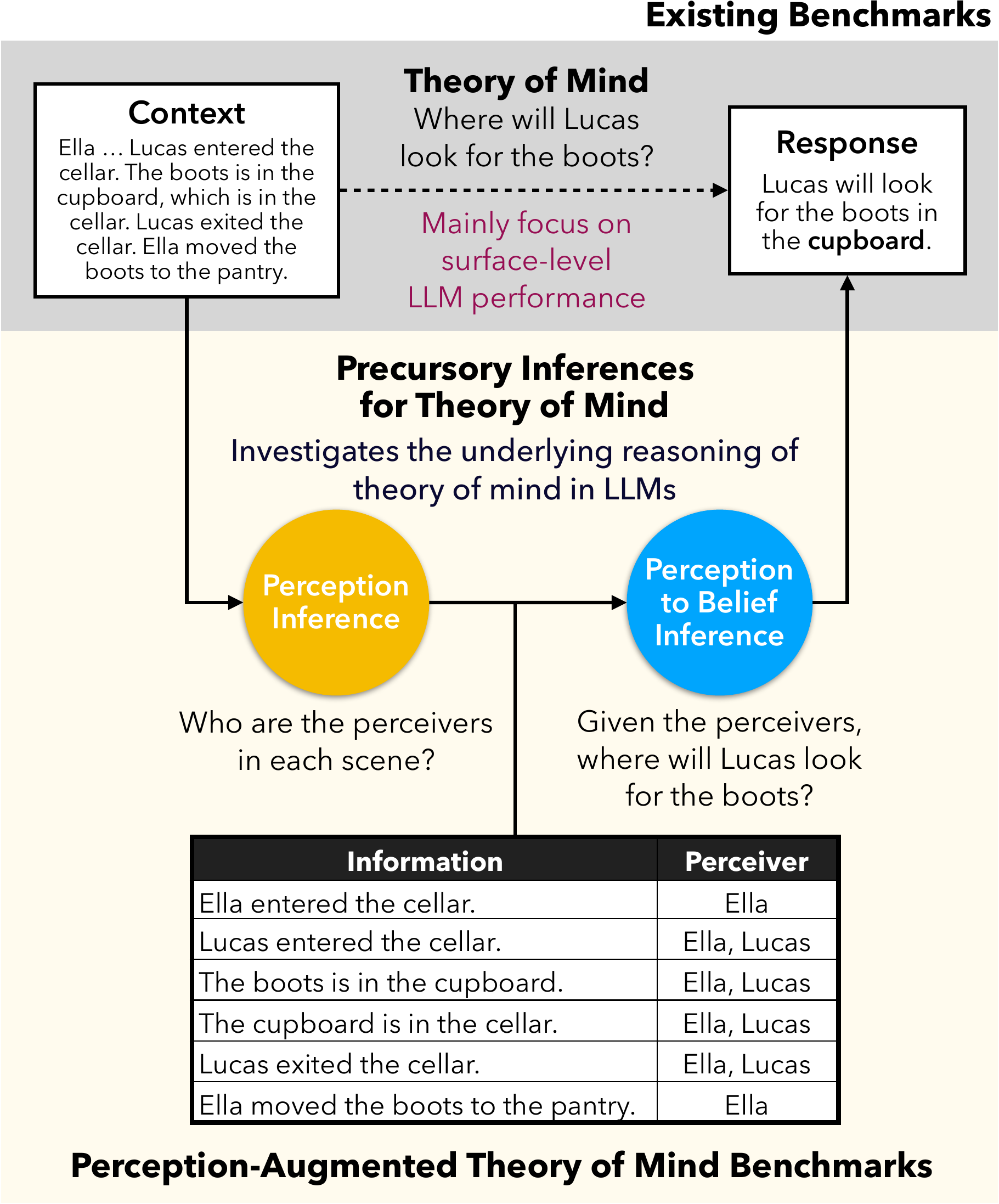}
  \caption{
  Inspired by children's developmental trajectory for theory of mind (ToM), our perception-augmented ToM benchmarks test the two precursory inferences of ToM in LLMs in order to examine their underlying social reasoning capabilities: (1) \textit{perception inference} and (2) \textit{perception-to-belief inference} (\S \ref{sec:method_benchmark}). 
  }
  \label{fig:overview}
\end{figure}

However, there are clear limitations to understanding the gaps in LLMs' underlying ToM abilities based on the evaluation results of existing ToM benchmarks, which only focus on the accuracy of the models' responses to ToM questions \citep{ma-etal-2023-towards-holistic}.
Although some studies have conducted error analysis based on model responses~\citep{ma-etal-2023-tomchallenges, wu-etal-2023-hi}, they rely on qualitative analysis via human inspection.

Psychology literature describes precursory steps to ToM development: \textit{perception inference}~\citep{rakoczy2022foundations} and \textit{perception-to-belief inference}---understanding that `\textit{seeing leads to knowing}'~\citep{pratt1990young, baron-cohen1994the}. 
These capabilities can be defined in the scenario shown in Figure~\ref{fig:overview}. We refer to the ability to infer others' perceptions (e.g., \textit{Who are the perceivers of each scene?}'') as \textit{perception inference} and the process of deducing others' beliefs from their perceptions (e.g., \textit{Given the perceivers of each scene, where will Lucas look for the boots?}'') as \textit{perception-to-belief inference}.

Inspired by the human developmental stages for ToM, we evaluate the key precursory inference steps of ToM in LLMs.
First, we extend the two representative ToM benchmarks, ToMi~\citep{le-etal-2019-revisiting} and FANToM~\citep{kim-etal-2023-fantom}, by annotating characters' perceptions about each piece of information from the input context. Figure~\ref{fig:overview} illustrates an example of our annotations and tasks on ToMi.
Second, using our new benchmarks, we evaluate eight state-of-the-art LLMs and find that models perform generally well in \textit{perception inference} but perform poorly in the \textit{perception-to-belief inference} task (\S \ref{subsec:results_p_inference} and \S \ref{subsec:results_ptob_inference}).
We also find that LLMs have weak \textit{inhibitory control} when inferring beliefs -- i.e., the capability of suppressing irrelevant information (\S \ref{subsec:inhibitory_control}).

Based on these findings, we propose PercepToM, a novel framework to enhance the ToM in LLMs by leveraging their perception inference capability. 
PercepToM first guides LLMs to infer the characters’ perceptions from an input context. 
Then, it aids LLMs in perception-to-belief inference through the \textit{perspective context extraction} step, which isolates the context perceived by the target character with a simple string-matching algorithm. Finally, LLMs answer to the ToM questions given the isolated context.
This approach leads to improved performance on both ToMi and FANToM, particularly on the false belief scenarios (\S \ref{subsec:result-perceptom}).

Our contributions are as follows.
First, we construct perception-augmented ToM benchmarks which enable the evaluation of the two precursory inferences for ToM in LLMs (\S\ref{sec:method_benchmark}): \textit{perception inference} and \textit{perception-to-belief inference}.
Second, using these benchmarks, we show that current LLMs are good at inferring the perceptions of others but struggle to infer beliefs from the perceptual information (\S\ref{subsec:results_p_inference}, \S \ref{subsec:results_ptob_inference}, and \S \ref{subsec:inhibitory_control}).
Lastly, we introduce the PercepToM framework to improve LLMs' ToM reasoning by leveraging their strong \textit{perception inference} while supplementing their \textit{perception-to-belief inference} (\S\ref{sec:method_percptom}).
Our method enhances LLMs' performance on ToMi and FANToM, especially on the false belief scenarios (\S\ref{subsec:result-perceptom}).

\begin{figure*}[t!]
  \includegraphics[width=\textwidth]{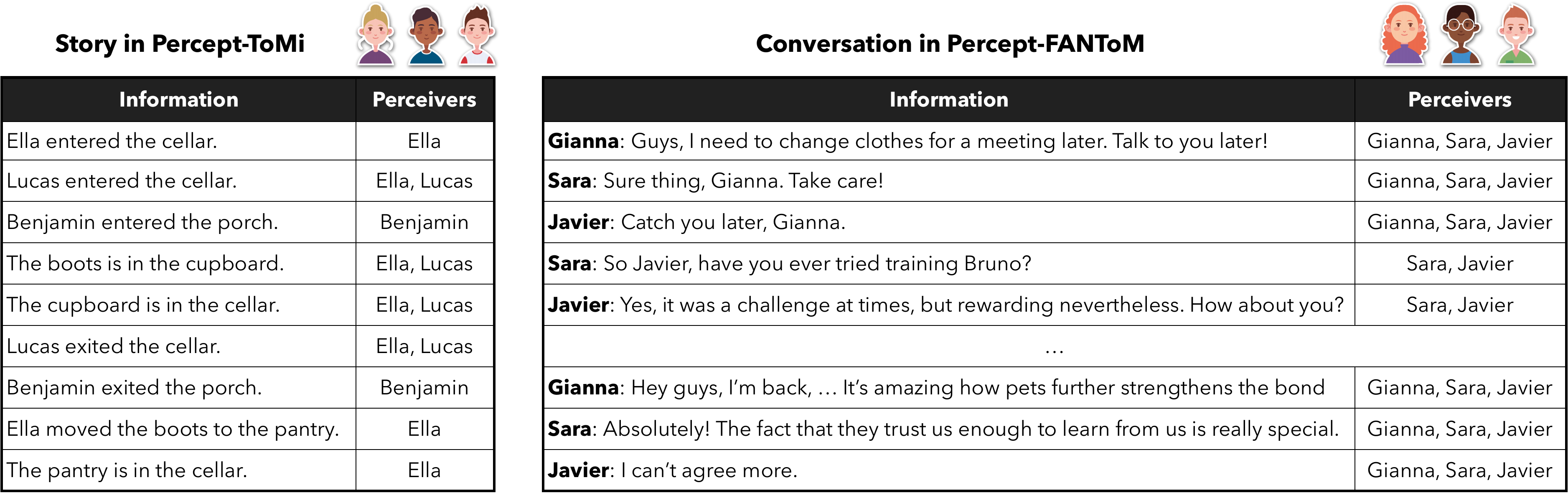}
  \caption{
  Example data in Percept-ToMi and Percept-FANToM.
  For each context, the perceivers of every scene description or utterance are annotated automatically (Percept-ToMi) and manually (Percept-FANToM).
  }
  \vspace{-0.2cm}
  \label{fig:annotation}
\end{figure*}

\section{Augmenting Perceptions on\newline Theory of Mind Benchmarks}\label{sec:method_benchmark}

We construct perception-augmented theory of mind (ToM) benchmarks to evaluate two essential cornerstones for ToM in LLMs: (1) \textit{perception inference} and (2) \textit{perception-to-belief inference} capabilities.

\subsection{Perception Inference and\newline Perception-to-Belief Inference}
\label{subsec:precursory}
The precursory inferences for ToM \citep{rakoczy2022foundations} can be understood through the Sally-Anne test, a widely recognized psychological assessment for evaluating ToM~\citep{Baron-Cohen1985-BARDTA}. In this test, Sally initially observes a marble in a box but does not witness Anne moving the marble to a basket after she leaves the room. 

Current ToM benchmarks predominantly assess LLMs based on their surface-level performance on ToM questions (e.g., ``\textit{Where will Sally look for the marble when she returns?}''), leaving their underlying inference capabilities underexplored. To address this gap, drawing from psychology literature, we refer to the ability to infer others' perceptions (e.g., ``\textit{Did Sally see Anne moving marble to the basket?}'') as \textit{perception inference}.  Additionally, we define the process of deducing beliefs from perceptual information (e.g., ``\textit{Sally did not see the marble being moved. Where will she look for the marble when she returns?}'') as \textit{perception-to-belief inference}.

To further investigate these inferences, we construct Percept-ToMi and Percept-FANToM by annotating each character's perception of information within the context of the two benchmark datasets ToMi~\citep{le-etal-2019-revisiting} and FANToM~\citep{kim-etal-2023-fantom}. Annotation examples are presented in Figure~\ref{fig:annotation}.

\subsection{The Source Theory of Mind Benchmarks}

\paragraph{ToMi \citep{le-etal-2019-revisiting}}
We include ToMi, one of the most widely used ToM benchmarks for reading comprehension tasks. 
The contexts in ToMi feature narrative scene descriptions, assuming characters acquire information by visual perception.
In each story, several characters are present in a room along with an object.
The story implicitly presumes that the characters can observe all objects and events taking place within the room.
There are four ToM question types in ToMi for a given story: first-order true/false beliefs, and second-order true/false beliefs.
In the true belief scenario, all characters observe everything happening in the room, ensuring that they share identical access to the information. However, in the false belief scenario, a character leaves the room, and then another character moves the object from one container to another, resulting in information asymmetry about the same object.

\paragraph{FANToM \citep{kim-etal-2023-fantom}}
This recent benchmark reveals a significant performance gap between humans and state-of-the-art LLMs.
It consists of multi-party conversations, assuming information transfer through both visual and auditory perceptions.
The information asymmetry occurs as some of the characters leave or join the conversation.
When a character is absent, the remaining participants share information exclusively among themselves.
FANToM also includes true belief scenarios where the absent character gets informed about the conversation upon rejoining the group.

\subsection{Perception-Augmented ToM Benchmarks}
\label{subsec:perception-tom}

\paragraph{Percept-ToMi}
To construct Percept-ToMi, we sample 150 story-question pairs for each of the four ToM question types in ToMi\footnote{We use the \textit{Fixed and Disambiguated ToMi} constructed by \citet{sclar-etal-2023-minding}, where sentences are inserted to disambiguate the location of containers in the story, and some mislabeled questions are corrected.}: first-order true/false beliefs, and second-order true/false beliefs.
We automatically annotate perceivers of the scenes in ToMi using SymbolicToM~\citep{sclar-etal-2023-minding} and manually verify the samples.
SymbolicToM tracks the witnesses of each scene by maintaining a graphical representation of the true world state, allowing us to obtain the list of perceivers for each scene from its output. 
However, upon verifying 50 samples of the SymbolicToM output, we identify two types of errors in the perceiver annotations and correct them across our entire dataset. The details of this verification and correction of perceiver annotations are explained in Appendix~\ref{appendix:percept-tomi}.

\paragraph{Percept-FANToM}
To build Percept-FANToM, we use all of the short conversations in FANToM, but exclude those that cause errors in our perception annotation format.
We assume that a character is the perceiver of all utterances that occur from the time they join the conversation until the time they leave. 
After two of the authors confirmed the criteria for determining the joining and leaving times (Appendix~\ref{appendix:percept-fantom}), each data point was manually annotated by one of the authors. They followed the annotation criteria mechanistically, ensuring no subjectivity was involved. Additionally, the authors randomly selected 20 samples to check for any discrepancies in their annotations. The results confirmed that all annotations were consistent between both authors.
Based on the annotations of characters' joining and leaving times, perceivers of each utterance are automatically mapped.
The statistics of our perception-augmented ToM benchmarks and the source benchmarks are shown in Appendix~\ref{appendix:data_stat}.

\subsection{Task and Evaluation}\label{sec:method_evaluation}
We measure the performance of (1) \textit{perception inference} and (2) \textit{perception-to-belief inference} in both false belief and true belief scenarios.

\paragraph{(1) Perception Inference}
\label{sec:perception_inference_capability}
In order to evaluate the perception inference capability of LLMs, we prompt the models to track characters' perception of each unit of information in the input context. 
Specifically, we require the models to respond in the format of a JSON array, which consists of JSON objects containing a unit of information from the context as a key and the perceivers of the information as a value.\footnote{We structure the perception inference results in JSON to leverage its parsability and interpretability. Also, recent works use JSON format to improve language model generation quality~\citep{zhou2023far, OpenAIJSONMode}.}
We use individual sentences and utterances as the units of information for ToMi and FANToM, respectively.
To ensure the generated answers are in the correct format, we provide an example format of the JSON array using a dummy sentence that does not appear in the datasets. The example input prompt is in Appendix~\ref{appendix:example_prompt_pi}.

\paragraph{(2) Perception-to-Belief Inference}
To evaluate the perception-to-belief inference capability of the models, we provide them with a ground truth perception inference result and then query ToM questions from the original benchmarks.
The ground truth perception inference result is provided in the same JSON array format we use to evaluate the perception inference capability of LLMs.
The example and detailed explanation of the input prompt can be found in Appendix~\ref{appendix:example_prompt_p2bi}.

\section{Precursory Inferences of ToM in LLMs}\label{sec:tom_evaluation}

\subsection{Experimental Setup}\label{subsec:tom_eval_experiments}

We analyze the \textit{perception inference} and \textit{perception-to-belief inference} (\S \ref{subsec:precursory}) performances of LLMs on Percept-ToMi and Percept-FANToM (\S \ref{subsec:perception-tom}) with the following metrics and models.

\paragraph{Perception Inference} 
To evaluate the model-generated perception inference results, we calculate accuracy for a given input context based on the ratio of information units where the model accurately identifies the perceivers. The final \textit{perception inference accuracy} for a dataset is obtained by averaging the accuracies across all contexts in the dataset. In Percept-ToMi, accuracy is calculated across the stories, with each story paired with a single ToM question.
In Percept-FANToM, since multiple questions share a single context, we calculate accuracy across these contexts.

\paragraph{Perception-to-Belief Inference and ToM}
\label{par:ptob_tom_metric}
We evaluate the perception-to-belief inference and ToM performance of LLMs using the original questions and answers from ToMi~\citep{le-etal-2019-revisiting} and FANToM~\citep{kim-etal-2023-fantom}.
For ToMi, we measure accuracy by the ratio of correctly answered questions among all story-question pairs.
Note that we do not use the \textit{joint accuracy} proposed in the original ToMi, where a story is counted as correctly answered only if all questions about the story are answered correctly. This is because many of the stories in the Fixed and Disambiguated ToMi~\citep{sclar-etal-2023-minding} do not include all six question types of ToMi.
For Percept-FANToM, we report the \textit{set:ALL
} score, which requires the model to correctly answer five types of ToM questions\footnote{\beliefChoiceQ, \answerabilityListQ, \infoaccessListQ, \answerabilityYNQ, \infoaccessYNQ} for the same piece of information within a conversation.

\paragraph{Correlation between LLM's ToM Performance and Precursory Inference Performance}
To analyze the relationship between LLMs' ToM capability and their performance on perception-related ToM precursor tasks (i.e., perception inference and perception-to-belief inference), we measure the Pearson correlation coefficient between models' performances on ToM and each of these two tasks.

\paragraph{Target Models}
We examine eight state-of-the-art LLMs: GPT-3.5 Turbo (\texttt{gpt-3.5-turbo-1106}), GPT-4 Turbo (\texttt{gpt-4-turbo-1106-preview}), GPT-4o (\texttt{gpt-4o-2024-05-13})\footnote{\url{https://platform.openai.com/docs/models/overview}}, Claude 3 (Haiku and Sonnet)\footnote{\url{https://www.anthropic.com/product}}, Gemini 1.0 Pro~\citep{geminiteam2024gemini}, Llama-3 70B Instruct~\citep{llama3modelcard}, and Mixtral 8x22B Instruct ~\citep{jiang2024mixtral} on Percept-ToMi and Percept-FANToM (\S \ref{subsec:perception-tom}).

\begin{figure*}[th!]
  \includegraphics[width=\linewidth]{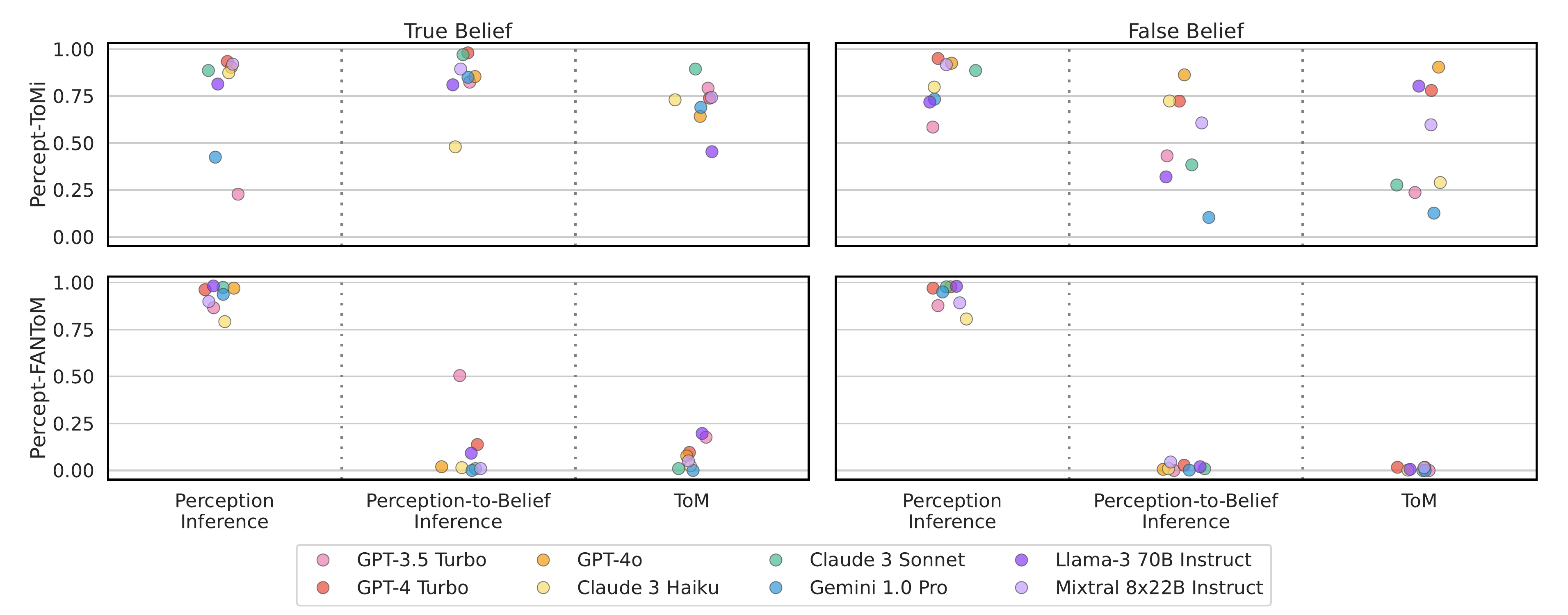}
  \caption{Perception inference, perception-to-belief inference, and ToM performances of LLMs in true and false belief scenarios of Percept-ToMi and Percept-FANToM. Although the models exhibit similar accuracy in perception inference across both scenarios, their performance in perception-to-belief inference and ToM scenarios varies significantly.}
  \label{fig:catplot_cap}
\end{figure*}

\begin{figure*}[t]
    \includegraphics[width=\linewidth]{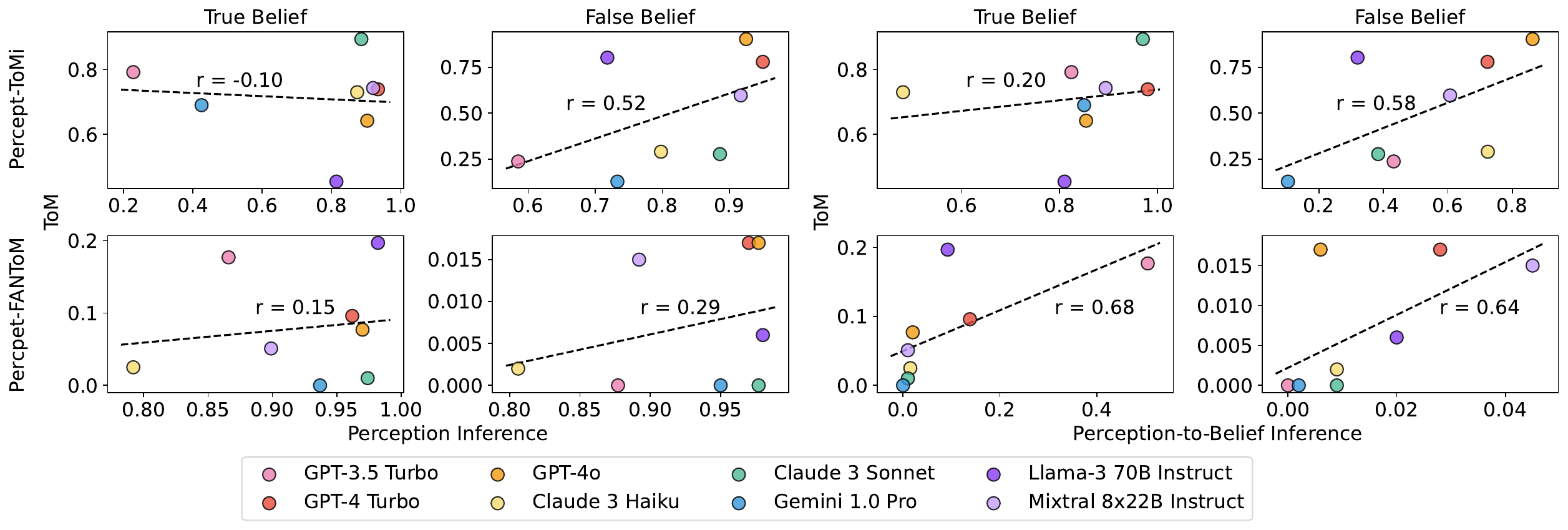}
    \caption{
    Pearson correlation of LLMs' ToM performance with perception inference (left) and perception-to-belief inference (right) performances. 
    ToM performance shows a positive correlation with perception-to-belief inference performance but exhibits a weak or no correlation with perception inference performance.
    }
    \label{fig:tom_perc_correlation}
\end{figure*}

\subsection{Perception Inference}
\label{subsec:results_p_inference}

\paragraph{LLMs generally perform well on perception inference across datasets and scenarios.}
As shown in Figure~\ref{fig:catplot_cap}, most LLMs exhibit high accuracy on perception inference in both Percept-ToMi and Percept-FANToM. The models' average perception inference accuracy is 0.781 on Percept-ToMi and 0.926 on Percept-FANToM. 
Also, they exhibit negligible differences in the accuracy between the true belief and false belief scenarios. 
In ToMi, all models except for GPT-3.5 Turbo and Gemini 1.0 Pro exhibit a gap of less than 0.1 accuracy between the two scenarios.
In FANToM, the accuracy gaps between the two scenarios in all models are no greater than 0.014.
This result contrasts with the models' large performance gap in the two scenarios on ToM questions, suggesting that their limited ToM performance in false belief scenarios is not due to the lack of perception inference capability.
Detailed results are in Appendix~\ref{appendix:catplot_cap_score}.

\paragraph{The perception inference and ToM performance do not show a strong correlation.}
\label{para:perc-scenario_comparison}

Especially in the ToMi's true belief scenarios, the two performances exhibit a near-zero correlation (Figure \ref{fig:tom_perc_correlation}).
Although moderate correlations appear in other scenarios, the correlation coefficients are not statistically significant.
These results imply that LLMs' perception inference capability is not directly linked to their ToM performance. 
This contrasts with human adults, where ToM is strictly dependent on perception inference.

\subsection{Perception-to-Belief Inference}
\label{subsec:results_ptob_inference}

\paragraph{LLMs struggle with perception-to-belief inference.}
Surprisingly, although the ground truth perception information for all characters is provided in this task, models still underperform in false belief scenarios compared to true belief scenarios (see Figure \ref{fig:catplot_cap}).
This trend is consistent with their ToM performance.
Moreover, their performances on the perception-to-belief inference task are mostly similar to their ToM performances in all scenarios except for the ToMi true belief scenario.
The fact that the LLMs hardly benefit from the additional character perception information, which should serve as significant hints for solving ToM questions, suggests that they have limited capability to infer beliefs from perceptions.
The exact performances of models are in Appendix~\ref{appendix:catplot_cap_score}.

\paragraph{The perception-to-belief inference and ToM performance exhibit a positive correlation.}
This is consistent across all datasets and scenarios (Figure~\ref{fig:tom_perc_correlation}). 
Notably, in FANToM, models exhibit a high correlation between the two performances ($r>0.6$). 
This correlation likely arises because the two tasks use the same questions.
However, since LLMs are showing similar performances in both tasks, we can see that they are not fully leveraging the ground truth perception information in the perception-to-belief inference task.

\section{PercepToM: Grounding ToM Reasoning on Perception}
\label{sec:method_percptom}

\subsection{Framework}
\begin{figure}[t]
  \includegraphics[width=\columnwidth]{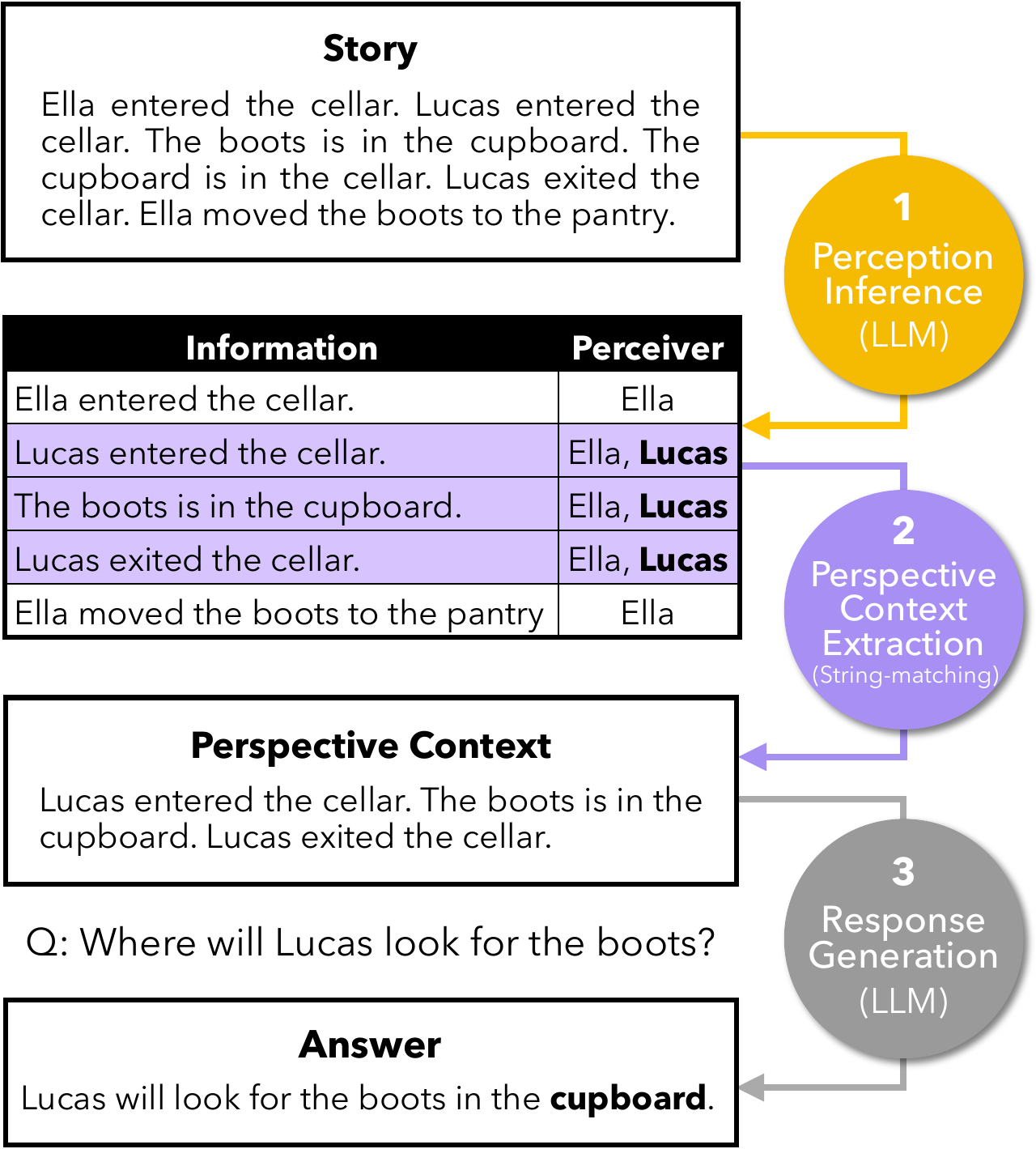}
  \caption{
  An overview of our PercepToM framework, which enhances LLMs' ToM reasoning by (1) instructing LLMs to infer the perceivers of each information in the context; (2) aiding their perception-to-belief inference through the \textit{perspective context extraction} step, which isolates the context perceived by the target character; and (3) allowing LLMs to generate responses to ToM questions based on this perspective context.
  }
  \label{fig:percepToM}
\end{figure}

According to our experiment results, LLMs perform adequately well in both true and false belief scenarios on perception inference, while they underperform in perception-to-belief inference (\S\ref{sec:tom_evaluation}).
Based on these findings, we propose PercepToM, a framework for improving LLM's ToM reasoning by grounding it in perception information.
PecepToM leverages LLM's strong perception inference capabilities while enhancing its perception-to-belief inference with a simple string-matching rule.
PercepToM consists of the following steps as illustrated in Figure~\ref{fig:percepToM}:

\begin{enumerate}[leftmargin=0.45cm]
    \item \textbf{Perception Inference}: The LLM infers which characters perceived each unit of information in the context (e.g., a scene description or an utterance).
    \item \textbf{Perspective Context Extraction}: Based on the perception inference result from the LLM, PercepToM extracts the \textit{perspective context} --- i.e., the subset of the input context identified by the LLM as perceived by the target character. This process is conducted by a simple string-matching procedure.
    \item \textbf{Response Generation}: Given the perspective context of the target character, the LLM answers the ToM question.
\end{enumerate}

If the model correctly performs perception inference, the perspective context will only include what the target character has perceived from the original context -- that is, what they believe to be true, based on the principle of rational belief \citep{Baker2011-xt}.
When given this isolated context along with the ToM question, the scenario becomes a simple true belief scenario, wherein the LLM has access to the same information as the target character (i.e., information symmetry).

SymbolicToM~\citep{sclar-etal-2023-minding} also helps LLM's ToM reasoning by providing only the context included in the target character's belief state graph to the model.
However, constructing the belief graph in SymbolicToM requires manually crafted algorithms tailored to different types of input.
In contrast, PercepToM avoids this requirement by leveraging LLM's perception inference capabilities, which can handle more diverse and complicated contexts, thereby achieving significantly improved generalizability.
The example input and output of each step of the algorithm are provided in Appendix~\ref{appendix:example_io_perceptom}.

\subsection{Experimental Setup}

\paragraph{Datasets and Metrics}
We evaluate PercepToM and other baseline models on Fixed and Disambiguated ToMi~\citep{le-etal-2019-revisiting, sclar-etal-2023-minding} and FANToM~\citep{kim-etal-2023-fantom}.
As evaluation metrics, we use the ratio of correctly answered story-question pairs for ToMi and the \textit{set:ALL} score for FANToM, as employed in the ToM performance evaluation in the previous section (\S\ref{par:ptob_tom_metric}).

\paragraph{Baselines}
We compare Vanilla, Chain-of-Thought \citep[CoT;][]{wei2022chain}, and System 2 Attention \citep[S2A;][]{weston20232} with PercepToM.
Vanilla involves LLM directly answering questions based on the given context, while CoT adds the prompt \textit{``Let's think step by step.''} to help the model answer ToM questions. S2A improves the reasoning of LLMs by prompting them to extract only the relevant part of the input context before yielding a final response. 
By using S2A as a baseline, we compare the effectiveness of the perspective context of PercepToM with the relevant context extracted by LLMs using S2A.
We also compare the performance of PercepToM to that of SymbolicToM \citep{sclar-etal-2023-minding} on ToMi.
However, we do not extend this comparison to FANToM, as applying SymbolicToM to input formats other than ToMi is not trivial, given that it is specifically tailored to ToMi's input format.

\paragraph{Target Models}
Since PercepToM leverages the perception reasoning capability of LLMs, we choose models that show reasonable performance on the perception inference task.
Specifically, among the eight models, we exclude the bottom two in terms of perception inference accuracy on Percept-FANToM and Percept-ToMi, which are GPT-3.5 Turbo, Claude 3 Haiku, and Gemini 1.0 Pro. 
As a result, we apply our PercepToM framework to GPT-4 Turbo, GPT-4o, Claude 3 Sonnet, Llama-3 70B Instruct, and Mixtral 8x22B.

\subsection{Results}
\label{subsec:result-perceptom}

Table~\ref{tab:perceptom_result} shows that the PercepToM improves overall ToM performance when applied to different LLMs on ToMi and FANToM.
Remarkably, GPT-4 Turbo achieves 1.0, a perfect score, on the false belief scenario in ToMi.
PercepToM generally outperforms CoT and S2A, suggesting that the perspective context extraction, grounded in LLMs' perception inference results, is more effective than either CoT reasoning or relevant context extraction in S2A at enhancing LLMs' ToM reasoning.

PercepToM's performance improvement is more pronounced in false belief scenarios than in true belief scenarios, likely because only minor parts of the contexts are filtered out during the perspective context extraction in the latter.
In the false belief scenario of FANToM, which is recognized as the most complex task, all LLMs equipped with PercepToM achieve the highest performance by a large margin.
For instance, Llama-3 70B Instruct achieves 0.147 when its vanilla performance is close to 0.

\begin{table}[t]
\begin{center}
\resizebox{\columnwidth}{!}{%
\begin{tabular}{@{}cccccc@{}}
\toprule
\multirow{2}{*}{Model} &
  \multirow{2}{*}{Method} &
  \multicolumn{2}{c}{ToMi} &
  \multicolumn{2}{c}{FANToM} \\ \cmidrule(l){3-6} 
 &
   &
  \begin{tabular}[c]{@{}c@{}}True\\ Belief\end{tabular} &
  \begin{tabular}[c]{@{}c@{}}False\\ Belief\end{tabular} &
  \begin{tabular}[c]{@{}c@{}}True\\ Belief\end{tabular} &
  \begin{tabular}[c]{@{}c@{}}False\\ Belief\end{tabular} \\ \midrule
\multirow{4}{*}{\begin{tabular}[c]{@{}c@{}}GPT-4 \\ Turbo\end{tabular}} &
  Vanilla &
  0.739 &
  0.780 &
  0.096 &
  0.017 \\
 &
  CoT &
  0.700 &
  0.930 &
  0.066 &
  0.079 \\
 &
  S2A &
  0.682 &
  0.727 &
  0.015 &
  0.019 \\ \cmidrule(l){2-6} 
 &
  PercepToM &
  \textbf{0.824} &
  \textbf{1.000} &
  \textbf{0.162} &
  \textbf{0.306} \\ \midrule
\multirow{4}{*}{GPT-4o} &
  Vanilla &
  0.642 &
  0.904 &
  0.077 &
  0.017 \\
 &
  CoT &
  \textbf{0.734} &
  \textbf{0.987} &
  \textbf{0.153} &
  0.241 \\
 &
  S2A &
  0.532 &
  0.933 &
  0.000 &
  0.006 \\ \cmidrule(l){2-6} 
 &
  PercepToM &
  0.659 &
  0.915 &
  0.117 &
  \textbf{0.566} \\ \midrule
\multirow{4}{*}{\begin{tabular}[c]{@{}c@{}}Claude 3 \\ Sonnet\end{tabular}} &
  Vanilla &
  0.894 &
  0.277 &
  0.010 &
  0.000 \\
 &
  CoT &
  0.610 &
  0.880 &
  0.005 &
  0.000 \\
 &
  S2A &
  0.870 &
  0.354 &
  0.000 &
  0.000 \\ \cmidrule(l){2-6} 
 &
  PercepToM &
  \textbf{0.963} &
  \textbf{0.937} &
  \textbf{0.035} &
  \textbf{0.066} \\ \midrule
\multirow{4}{*}{\begin{tabular}[c]{@{}c@{}}Llama-3\\ 70B Inst.\end{tabular}} &
  Vanilla &
  0.454 &
  0.803 &
  0.197 &
  0.006 \\
 &
  CoT &
  0.644 &
  \textbf{0.900} &
  0.081 &
  0.046 \\
 &
  S2A &
  0.410 &
  0.894 &
  0.020 &
  0.037 \\ \cmidrule(l){2-6} 
 &
  PercepToM &
  \textbf{0.713} &
  0.744 &
  \textbf{0.242} &
  \textbf{0.147} \\ \midrule
\multirow{4}{*}{\begin{tabular}[c]{@{}c@{}}Mixtral \\ 8x22B Inst.\end{tabular}} &
  Vanilla &
  0.743 &
  0.597 &
  0.051 &
  0.015 \\
 &
  CoT &
  0.567 &
  0.630 &
  0.010 &
  0.007 \\
 &
  S2A &
  \textbf{0.750} &
  0.357 &
  0.020 &
  0.007 \\ \cmidrule(l){2-6} 
 &
  PercepToM &
  0.727 &
  \textbf{0.964} &
  \textbf{0.217} &
  \textbf{0.035} \\ \bottomrule
\end{tabular}%
}
\caption{PercepToM outperforms the baseline models in most of the scenarios on ToMi and FANToM. Bold indicates the best performance within each language model and scenario (true belief or false belief). Performance comparison between PercepToM and SymbolicToM on ToMi can be found in Appendix~\ref{sec:perceptom-symbolictom-comparison}.}
\label{tab:perceptom_result}
\end{center}
\end{table}

The performance of PercepToM is also compared with that of SymbolicToM~\citep{sclar-etal-2023-minding} on ToMi (Appendix~\ref{sec:perceptom-symbolictom-comparison}).
PercepToM performs comparably to SymbolicToM in false belief scenarios across most LLMs.
However, in true belief scenarios, SymbolicToM consistently outperforms both PercepToM and PercepToM+Oracle.
We speculate that this performance gap arises because SymbolicToM rephrases the ToM questions into simpler reality questions.
For example, the ToM question ``\textit{Where will Bob look for the celery?}'' gets rephrased into ``\textit{Where is the celery?}''
In contrast, PercepToM addresses the ToM questions as is.

\subsection{Impact of Irrelevant Information on Perception-to-Belief Inference}
\label{subsec:inhibitory_control}
We conduct an ablation study on perspective context extraction in PercepToM to demonstrate the impact of irrelevant information on LLMs' perception-to-belief inference.
To remove the impact of LLMs' perception inference accuracy, we compare their performance on perception-to-belief inference with that of PercepToM+Oracle.
Both setups have access to the ground truth perception inference information; however, the PercepToM+Oracle includes the perspective context extraction step, while the perception-to-belief inference setup does not.

As Table~\ref{tab:ptob_oracle} shows, models perform significantly better in the PercepToM+Oracle setup than the perception-to-belief inference setup in most scenarios.
This suggests that in the perception-to-belief inference setting, despite the presence of the ground truth perception inference information -- which should be a substantial hint -- within the context, the inclusion of irrelevant information (e.g., the perception of non-target characters and the context not perceived by the target character) results in suboptimal performance in LLMs.
Therefore, we can see LLMs struggle to effectively suppress irrelevant information.
This capability, coined `\textit{inhibitory control}' in cognitive science, involves the ability to block out irrelevant stimuli while following a specific cognitive objective \citep{rothbart1985temperament}.
Inhibitory control is known to be closely linked to ToM and is considered a crucial component for developing ToM \citep{Carlson2001-mn, Carlson2002-ef}.

\section{Related Work}
\label{sec:related_work}

\paragraph{Benchmarks for LLM's Theory of Mind}
There has been a growing number of benchmarks aimed to evaluate LLM's theory of mind (ToM), including ToMi~\citep{le-etal-2019-revisiting}, FANToM~\citep{kim-etal-2023-fantom}, BigToM~\citep{gandhi2023understanding}, HI-TOM~\citep{wu-etal-2023-hi}, ToMChallenges~\citep{ma-etal-2023-tomchallenges}, Adv-CSFB~\citep{shapira-etal-2024-clever}, and OpenToM \citep{xu2024opentom}.
Most of them adopt the false belief test~\citep{WIMMER1983103}, a famous psychology test developed to assess human ToM capabilities. 
These benchmarks present scenarios involving a character who holds a false belief about a situation (e.g., not knowing something has changed).
Models are then asked to predict the character's thoughts or actions based on the false belief in the scenario.
Many benchmarks also include control scenarios where characters do not hold false belief (i.e., true belief scenarios) -- situations where their belief about the world state matches the actual state~\citep{le-etal-2019-revisiting, kim-etal-2023-fantom, gandhi2023understanding, shapira-etal-2024-clever}.

Unlike existing benchmarks that primarily measure performance on (downstream) ToM questions themselves, we aim to inquire into the underlying reasoning abilities of LLM's theory of mind by examining the precursor of ToM: the concept of \textit{seeing leads to knowing} \citep{baron-cohen1994the,pratt1990young}.
We expand existing datasets to identify the perception inference and perception-to-belief inference capabilities, which are essential for ToM reasoning.

\begin{table}[]
\resizebox{\columnwidth}{!}{%
\begin{tabular}{@{}cccccc@{}}
\toprule
\multirow{2}{*}{Model} &
  \multirow{2}{*}{Method} &
  \multicolumn{2}{c}{ToMi} &
  \multicolumn{2}{c}{FANToM} \\ \cmidrule(l){3-6} 
 &
   &
  \begin{tabular}[c]{@{}c@{}}True\\ Belief\end{tabular} &
  \begin{tabular}[c]{@{}c@{}}False\\ Belief\end{tabular} &
  \begin{tabular}[c]{@{}c@{}}True\\ Belief\end{tabular} &
  \begin{tabular}[c]{@{}c@{}}False\\ Belief\end{tabular} \\ \midrule
\multirow{2}{*}{\begin{tabular}[c]{@{}c@{}}GPT-4\\ Turbo\end{tabular}} &
  \begin{tabular}[c]{@{}l@{}}Perception-\\ to-Belief\end{tabular} &
  \textbf{0.980} &
  0.723 &
  0.138 &
  0.028 \\
 &
  \begin{tabular}[c]{@{}l@{}}PercepToM\\+Oracle\end{tabular} &
  0.885 &
  \textbf{0.993} &
  \textbf{0.270} &
  \textbf{0.336} \\ \midrule
\multirow{2}{*}{GPT-4o} &
  \begin{tabular}[c]{@{}l@{}}Perception-\\ to-Belief\end{tabular} &
  \textbf{0.854} &
  0.863 &
  0.020 &
  0.006 \\
 &
  \begin{tabular}[c]{@{}l@{}}PercepToM\\+Oracle\end{tabular} &
  0.660 &
  \textbf{0.993} &
  \multicolumn{1}{l}{\textbf{0.102}} &
  \multicolumn{1}{l}{\textbf{0.571}} \\ \midrule
\multirow{2}{*}{\begin{tabular}[c]{@{}c@{}}Claude 3\\ Sonnet\end{tabular}} &
  \begin{tabular}[c]{@{}l@{}}Perception-\\ to-Belief\end{tabular} &
  0.970 &
  0.384 &
  0.010 &
  0.009 \\
 &
  \begin{tabular}[c]{@{}l@{}}PercepToM\\+Oracle\end{tabular} &
  \textbf{0.987} &
  \textbf{0.987} &
  \textbf{0.031} &
  \textbf{0.058} \\ \midrule
\multirow{2}{*}{\begin{tabular}[c]{@{}c@{}}Llama-3\\ 70B Inst.\end{tabular}} &
  \begin{tabular}[c]{@{}l@{}}Perception-\\ to-Belief\end{tabular} &
  \textbf{0.810} &
  0.320 &
  0.092 &
  0.020 \\
 &
  \begin{tabular}[c]{@{}l@{}}PercepToM\\+Oracle\end{tabular} &
  0.677 &
  \textbf{0.980} &
  \textbf{0.133} &
  \textbf{0.161} \\ \midrule
\multirow{2}{*}{\begin{tabular}[c]{@{}c@{}}Mixtral\\ 8x22B Inst.\end{tabular}} &
  \begin{tabular}[c]{@{}l@{}}Perception-\\ to-Belief\end{tabular} &
  \textbf{0.894} &
  0.607 &
  0.010 &
  \textbf{0.045} \\
 &
  \begin{tabular}[c]{@{}l@{}}PercepToM\\+Oracle\end{tabular} &
  0.757 &
  \textbf{0.970} &
  \textbf{0.224} &
  0.039 \\ \bottomrule
\end{tabular}%
}
\caption{LLMs perform significantly better in PercepToM+Oracle than perception-to-belief inference across most scenarios.
Since the only difference between the two tasks is the inclusion of a perspective context extraction step in PercepToM+Oracle, the result suggests that LLMs struggle to suppress irrelevant perception information when solving ToM questions.}
\label{tab:ptob_oracle}
\end{table}

\paragraph{Improving LLM's Theory of Mind}
Previous research has explored several methods to enhance LLM's ToM ability. SymbolicToM~\citep{sclar-etal-2023-minding} tracks multiple characters' beliefs using graphical representation to provide LLMs the context in the target character's point of view. However, the necessity to construct the belief state graph restricts its adaptability in complex scenarios involving diverse relationships and interactions between entities.
SimToM~\citep{wilf2023think} improves LLM's ToM ability through prompt tuning and highlights the significance of perspective-taking.
ToM-LM~\citep{tang2024tomlm} improves performance through LLM fine-tuning, while it requires additional training resources. 

While SymbolicToM and SimToM achieve high performance on ToMi, their algorithms are tailored for the ToMi structure.
For example, SimToM's prompts include ToMi-specific hints, such as specific lists of events a character should be aware of for ToMi and instructions for output to depend on a fixed pattern in BigToM stories.
As a result, SimToM is mainly tested on ToMi and a specific subset of BigToM questions that resemble those in ToMi.
Similarly, the graph construction pipeline in SymbolicToM is designed specifically for ToMi.
In contrast, our method, PercepToM, demonstrates flexibility and effectiveness when applied to data of varying formats, leading to significantly better generalizability compared to the existing methods.
Although we do not include these two baselines in our main experiments, we present performance comparisons between PercepToM and these methods in Appendices~\ref{sec:perceptom-symbolictom-comparison} and \ref{sec:perceptom-simtom-comparison}.

\section{Conclusion}\label{sec:conclusion}

Inspired by the psychology literature, we evaluated the precursory inferences for human theory of mind (ToM) in large language models (LLM), aiming to broaden our insight into their ToM capabilities. 
To this end, we constructed perception-augmented ToM benchmarks, Percept-ToMi and Percept-FANToM, by annotating character perceptions about the contexts.
Through evaluations and analyses of eight state-of-the-art LLMs, we found that they perform reasonably well in inferring others' perceptions but struggle with inferring others' beliefs based on that perceptual information.
Based on these findings, we proposed a new framework, PercepToM, to improve LLM's ToM reasoning.
Our framework leverages LLMs' strength in perception inference and enhances their perception-to-belief inference by extracting the relevant contexts.
We expect our work to provide insights and enable further in-depth studies into the extent of LLMs' ToM capabilities and targeted improvements in their weaknesses.

\section{Limitations}\label{sec:limitations}

In this paper, we conduct experiments using only two text-based ToM datasets. While ToM tests in psychology involve visual stimuli (e.g., puppets or image strips), our evaluation of ToM abilities relies on text, requiring the ability to read and understand language. As a result, our models must possess robust language comprehension abilities. Moving forward, we are considering expanding our research to include visual ToM and multimodal ToM evaluations, exploring beyond text-based LLMs.

We compare LLMs' ToM performances between true belief and false belief scenarios, but not those between the different orders of ToM questions (e.g., first-order and second-order).  
Since higher-order ToM requires more inference steps, it will also be interesting to examine the differences in model behavior and capability in solving different orders of ToM questions in future work.

We analyze the precursory inferences for ToM in state-of-the-art large language models (LLMs) that are trained with the full conventional pipeline -- i.e., pretraining, instruction tuning, and preference tuning. To understand whether LLMs follow developmental stages akin to human cognition, it is crucial to conduct experiments across the training phases of LLMs. This would include investigating at which stage LLM's social reasoning abilities emerge.
These assessments will help us understand how the models' development of social reasoning aligns with stages observed in human theory of mind (ToM).

\section{Societal and Ethical Considerations}

Our use of FANToM dataset is consistent with its intended use, which is evaluation.
We have adhered to the licenses of the benchmarks, ToMi and FANToM, in processing them to create our benchmarks, Percept-ToMi and Percept-FANToM.
We plan to make our benchmarks publicly available with the license of Attribution-Noncommercial 4.0 International (CC BY-NC 4.0), allowing the sharing and adapting of the material.

Although we are analyzing large language models' (LLM) theory of mind (ToM) capabilities and its perception-related precursors, we emphasize that we do not claim these LLMs have a mind or any form of subjective consciousness.
Our focus lies on improving the social reasoning capabilities of these models to help them interact better in real-world social situations.

\section*{Acknowledgment}

We thank Amazon for their gift in support of research on theory of mind. This work was supported by Institute of Information \& communications Technology Planning \& Evaluation (IITP) grant funded by the Korea government(MSIT) (No. RS-2024-00443251, Accurate and Safe Multimodal, Multilingual Personalized AI Tutors).

\bibliography{anthology,custom}

\clearpage
\appendix

\section{Details of Perception-Augmented ToM Benchmarks}
\subsection{Manual Verification of Perception Annotation in Percept-ToMi}
\label{appendix:percept-tomi}
SymbolicToM identifies the perceivers of a scene in ToMi deterministically based on a graphical representation of the world state constructed from a templated description of the scene. However, since it determines the perceiver of a scene as all entities in the same connected component based on the world graph, it produces two types of errors in the perceiver information, as shown in Table~\ref{tab:tomi-perceiver-correction}. By reviewing 50 samples of SymbolicToM output, we discover these error types and correct every occurrence of them throughout our dataset.

The perceiver of distractor sentences in ToMi, which describe a character’s opinion about an object, should be the character themselves, since they do not express their opinion to others in the scenario. However, the SymbolicToM-generated perceiver annotation includes other characters located in the same space. We therefore correct the perceivers for all such distractor sentences.
Another error in perceiver annotations occur in the sentences preceding the location-disambiguating sentence, which specifies object locations, where their perceivers are annotated with `none.' We align the perceiver annotations of these sentences with those of the subsequent location-disambiguating sentenc, since they are always paired and have the same perceivers.

\begin{table*}[]
\begin{center}
{\small
\begin{tabular}{@{}cccc@{}}
\toprule
Sentence Type & Information & \begin{tabular}[c]{@{}c@{}}SymbolicToM\\Output\end{tabular} & Final Annotation \\ \midrule
\multirow{2}{*}{Object Location} & The slacks is in the pantry.         & None                & Ella, Benjamin \\
                                 & The pantry is in the master bedroom. & Ella, Benjamin      & Ella, Benjamin \\ \midrule
Distractor                       & Olivia loves the skirt.              & Olivia, James, Lily & Olivia         \\ \bottomrule
\end{tabular}
}
\caption{The example perceiver annotations in ToMi corrected by manual verification.}
\label{tab:tomi-perceiver-correction}
\end{center}
\end{table*}

\subsection{Perception Annotation Criteria for Percept-FANToM}
\label{appendix:percept-fantom}
The following criteria are used to determine the joining and leaving times of a character in a conversation within the FANToM dataset.
\begin{itemize}[leftmargin=0.45cm]
    \item When a character joins a conversation is determined by the moment the character directly participates in the conversation. If a character enters with an utterance like ``you guys are having an interesting conversation,'' we consider him/her a perceiver from the moment he/she starts speaking, as the exact point when the character began listening is unclear.
    \item When a character leaves the conversation is determined by the final farewell utterance. Even if a character disappears mid-utterance (e.g., C: ``Bye, A. So, B, what do you think?''), the entire utterance is still considered as perceived by the departing character.
 \end{itemize}

\section {Dataset Statistics}
\label{appendix:data_stat}
Table~\ref{tab:data_stat} presents a comparison of data statistics between perception-augmented ToM benchmarks and their corresponding source benchmarks.

\begin{table*}[]
\begin{center}
{\small
\begin{tabular}{@{}cccccccccccc@{}}
\toprule
\multicolumn{2}{c}{\multirow{3}{*}{Datasets}} & \multicolumn{5}{c}{True Belief} & \multicolumn{5}{c}{False Belief} \\ \cmidrule(l){3-12} 
\multicolumn{2}{l}{} & \multirow{2}{*}{\# Ctx.} & \multirow{2}{*}{\# Q.} & \multicolumn{3}{c}{Avg. \# Q. per Ctx.} & \multirow{2}{*}{\# Ctx.} & \multirow{2}{*}{\# Q.} & \multicolumn{3}{c}{Avg. \# Q. per Ctx.} \\ \cmidrule(lr){5-7} \cmidrule(l){10-12} 
\multicolumn{2}{l}{} &  &  & P.I. & PtoB & ToM &  &  & P.I. & PtoB & ToM \\ \midrule
\multicolumn{2}{c}{\begin{tabular}[c]{@{}l@{}}Disambiguated ToMi - Test Set\end{tabular}} & 2793 & 2793 & - & - & 1.0 & 1210 & 1210 & - & - & 1.0 \\
\multicolumn{2}{c}{Percept-ToMi} & 300 & 1802 & 4.0 & 1.0 & 1.0 & 300 & 1804 & 4.0 & 1.0 & 1.0 \\ \midrule
\multicolumn{2}{c}{FANToM} & 402 & 3432 & - & - & 8.5 & 642 & 8530 & - & - & 13.3 \\
\multicolumn{2}{c}{Percept-FANToM} & 340 & 13161 & 24.2 & 7.3 & 7.3 & 539 & 25279 & 23.3 & 11.8 & 11.8 \\ \bottomrule
\end{tabular}%
}
\caption{Comparison of dataset statistics between the source datasets and our proposed datasets. Number of contexts, the total number of questions, and the average number of questions per context for each task are compared. Our benchmarks expand upon the sampled contexts from the source datasets by incorporating two precursory inference tasks for Theory of Mind (ToM)—perception inference and perception-to-belief inference.}
\label{tab:data_stat}
\end{center}
\end{table*}

\section{Prompt Examples}

This section introduces prompt examples to evaluate perception inference and perception-to-belief inference.

\subsection{Perception Inference}\label{appendix:example_prompt_pi}

The following two boxes are prompt examples using Percept-ToMi and Percept-FANToM, respectively. Some parts are omitted because of the space limit.

\begin{mdframed}
\scriptsize{\texttt{Story: Ella likes the suit. Ella entered the cellar. Lucas entered the cellar. Benjamin entered the porch. The boots is in the cupboard. The cupboard is in the cellar. Lucas exited the cellar. Benjamin exited the porch. Ella likes the sweatshirt. Lucas entered the porch. Ella moved the boots to the pantry. The pantry is in the cellar. \\ \\
Create a JSON array consisting of JSON objects. Each object should contain a sentence from the story and the perceivers of the scene described in that sentence. Assume that characters in the story can perceive every scene occurring in their location but not scenes occurring elsewhere. Also, include the actant of any action as a perceiver of that action.\\
Provide only a JSON array in the following format. Do not include any explanation.\\
}}
\scriptsize{\texttt{[\{"Noah exited the living room.": ["Noah", "Emma"]\},]}}
\end{mdframed}

\begin{mdframed}
\scriptsize{\texttt{Gianna: Guys, I've really enjoyed sharing our pet stories, but I need to excuse myself. I need to change clothes for a meeting later. Talk to you later!\\
Sara: Sure thing, Gianna. Take care!\\
Javier: Catch you later, Gianna.\\
Sara: So Javier, have you ever tried training Bruno?\\
Javier: Yes, I did actually. It was a challenge at times, but rewarding nevertheless. How about you? Did you try training Snowflake?\\
...\\
Gianna: Hey guys, I'm back, couldn't miss out on more pet stories. Speaking of teaching and training pets, it is amazing how that further strengthens the bond between us and our pets, right?\\
...\\
Create a JSON array consisting of JSON objects. Each object should include an utterance from the dialogue and the audience for that utterance. Assume that characters in the story can hear every utterance that occurs while they are involved in the dialogue, but not those that occur when they are absent. Also, ensure that the speaker of each utterance is included in the audience. Provide only the JSON array in the following format. Do not include any explanations. \\}}
\scriptsize{\texttt{[\{"Noah: Hi, Emma.": ["Noah", "Emma"]\},]}}
\end{mdframed}

\subsection{Perception-to-Belief Inference}\label{appendix:example_prompt_p2bi}

The following two boxes are prompt examples using Percept-ToMi and Percept-FANToM, respectively. Some parts are omitted because of the space limit.

\begin{mdframed}
\scriptsize{\texttt{Each JSON object in the following list contains the description of a consecutive scene in a story and its perceivers.\\ \\}}
\scriptsize{\texttt{[\{"Ella likes the suit": ["Ella"]\},\\
 \{"Ella entered the cellar": ["Ella"]\},\\
 \{"Lucas entered the cellar": ["Lucas", "Ella"]\},\\
 \{"Benjamin entered the porch": ["Benjamin"]\},\\
 \{"The boots is in the cupboard": ["Ella", "Lucas"]\},\\
 \{"The cupboard is in the cellar": ["Ella", "Lucas"]\},\\
 \{"Lucas exited the cellar": ["Lucas", "Ella"]\},\\
 \{"Benjamin exited the porch": ["Benjamin"]\},\\
 \{"Ella likes the sweatshirt": ["Ella"]\},\\
 \{"Lucas entered the porch": ["Lucas"]\},\\
 \{"Ella moved the boots to the pantry": ["Ella"]\},\\
 \{"The pantry is in the cellar": ["Ella"]\}]\\ \\
Question: Where will Lucas look for the boots? State the most detailed position possible. (e.g., in A in B) Answer in one sentence without explanation.\\
Answer: 
}}
\end{mdframed}

\begin{mdframed}
\scriptsize\texttt{{Each JSON object in the following list contains consecutive utterances in a dialogue and its audiences.\\ \\}}
\scriptsize\texttt{{[\{"Gianna: Guys, I've really enjoyed sharing our pet stories, but I need to excuse myself. I need to change clothes for a meeting later. Talk to you later!": ["Gianna", "Sara", "Javier"]\},\\
\{"Sara: Sure thing, Gianna. Take care!": ["Sara", "Gianna"]\},\\
\{"Javier: Catch you later, Gianna.": ["Javier", "Gianna"]\},\\
\{"Sara: So Javier, have you ever tried training Bruno?": ["Sara", "Javier"]\},\\
\{"Javier: Yes, I did actually. It was a challenge at times, but rewarding nevertheless. How about you? Did you try training Snowflake?": ["Javier", "Sara"]\},\\
...\\
\{"Gianna: Hey guys, I'm back, couldn't miss out on more pet stories. Speaking of teaching and training pets, it is amazing how that further strengthens the bond between us and our pets, right?": ["Gianna", "Sara", "Javier"]\},\\
...]\\ \\
Target: Who discussed their experiences training their pets, Bruno and Snowflake? \\
Question: Does Javier know the precise correct answer to this question? Answer yes or no. \\
Answer:
}}
\end{mdframed}

\section{Input and Output Examples of PercepToM Pipeline}\label{appendix:example_io_perceptom}

This section presents examples of input prompts and intermediate outputs of PercepToM steps. Note that PercepToM consists of three steps: perception inference, perspective context extraction, and reading comprehension.

First, the following two boxes are prompts for character perception inference on ToMi and FANToM, respectively.

\begin{mdframed}
\scriptsize{\texttt{Story: Ella likes the suit. Ella entered the cellar. Lucas entered the cellar. Benjamin entered the porch. The boots is in the cupboard. The cupboard is in the cellar. Lucas exited the cellar. Benjamin exited the porch. Ella likes the sweatshirt. Lucas entered the porch. Ella moved the boots to the pantry. The pantry is in the cellar. \\ \\
Create a JSON array consisting of JSON objects. Each object should contain a sentence from the story and the perceivers of the scene described in that sentence. Assume that characters in the story can perceive every scene occurring in their location but not scenes occurring elsewhere. Also, include the actant of any action as a perceiver of that action. \\
Provide only a JSON array in the following format. Do not include any explanation.\\}}
\scriptsize{\texttt{[\{"Noah exited the living room.": ["Noah", "Emma"]\},]}}
\end{mdframed}

\begin{mdframed}
\scriptsize{\texttt{Gianna: Guys, I've really enjoyed sharing our pet stories, but I need to excuse myself. I need to change clothes for a meeting later. Talk to you later!\\
Sara: Sure thing, Gianna. Take care!\\
Javier: Catch you later, Gianna.\\
Sara: So Javier, have you ever tried training Bruno?\\
Javier: Yes, I did actually. It was a challenge at times, but rewarding nevertheless. How about you? Did you try training Snowflake?\\
Sara: Oh gosh, trying to train a cat is a whole different ball game. But I did manage to teach her a few commands and tricks. She was quite an intelligent little furball.\\
Gianna: Hey guys, I'm back, couldn't miss out on more pet stories. Speaking of teaching and training pets, it is amazing how that further strengthens the bond between us and our pets, right?\\
...\\ \\
Create a JSON array consisting of JSON objects. Each object should contain a sentence from the story and the perceivers of the scene described in that sentence. Assume that characters in the story can perceive every scene occurring in their location but not scenes occurring elsewhere. Also, include the actant of any action as a perceiver of that action. Provide only a JSON array in the following format. Do not include any explanation. \\}}
\scriptsize{\texttt{[\{"Noah exited the living room.": ["Noah", "Emma"]\},]}}
\end{mdframed}

Here are the outputs of perception inference, which will be the input for perspective context extraction.

\begin{mdframed}
\scriptsize\begin{verbatim}[{"Ella likes the suit.": ["Ella"]},
 {"Ella entered the cellar.": ["Ella"]},
 {"Lucas entered the cellar.": ["Ella", "Lucas"]},
 {"Benjamin entered the porch.": ["Benjamin"]},
 {"The boots is in the cupboard.": ["Ella", "Lucas"]},
 {"The cupboard is in the cellar.": ["Ella", "Lucas"]},
 {"Lucas exited the cellar.": ["Lucas"]},
 {"Benjamin exited the porch.": ["Benjamin"]},
 {"Ella likes the sweatshirt.": ["Ella"]},
 {"Lucas entered the porch.": ["Lucas"]},
 {"Ella moved the boots to the pantry.": ["Ella"]},
 {"The pantry is in the cellar.": ["Ella"]}]
}\end{verbatim}
\end{mdframed}

\begin{mdframed}
\scriptsize\texttt{{[\{"Gianna: Guys, I've really enjoyed sharing our pet stories, but I need to excuse myself. I need to change clothes for a meeting later. Talk to you later!": ["Gianna", "Sara", "Javier"]\},\\
 \{"Sara: Sure thing, Gianna. Take care!": ["Sara", "Gianna"]\},\\
 \{"Javier: Catch you later, Gianna.": ["Javier", "Gianna"]\},\\
 \{"Sara: So Javier, have you ever tried training Bruno?": ["Sara", "Javier"]\},\\
 \{"Javier: Yes, I did actually. It was a challenge at times, but rewarding nevertheless. How about you? Did you try training Snowflake?": ["Javier", "Sara"]\},\\
 \{"Sara: Oh gosh, trying to train a cat is a whole different ball game. But I did manage to teach her a few commands and tricks. She was quite an intelligent little furball.": ["Sara", "Javier"]\},\\
 \{"Gianna: Hey guys, I'm back, couldn't miss out on more pet stories. Speaking of teaching and training pets, it is amazing how that further strengthens the bond between us and our pets, right?": ["Gianna", "Sara", "Javier"]\},
...]
}}
\end{mdframed}

The perspective context extraction selects the subset of context perceived by the target character. The outputs will be as follows:

\begin{mdframed}
\scriptsize{\texttt{Lucas entered the cellar. The boots is in the cupboard. The cupboard is in the cellar. Lucas exited the cellar. Lucas entered the porch.
}}
\end{mdframed}

\begin{mdframed}
\scriptsize\texttt{{Gianna: Guys, I've really enjoyed sharing our pet stories, but I need to excuse myself. I need to change clothes for a meeting later. Talk to you later!\\
Sara: Sure thing, Gianna. Take care!\\
Javier: Catch you later, Gianna.\\
Gianna: Hey guys, I'm back, couldn't miss out on more pet stories. Speaking of teaching and training pets, it is amazing how that further strengthens the bond between us and our pets, right?\\
...
}}
\end{mdframed}

\begin{table*}[]
\begin{center}
{\small
\resizebox{\textwidth}{!}{%
\begin{tabular}{@{}cccccccc@{}}
\toprule
 &
   &
  \multicolumn{3}{c}{True Belief} &
  \multicolumn{3}{c}{False Belief} \\ \cmidrule(l){3-8} 
\multirow{-2}{*}{Dataset} &
  \multirow{-2}{*}{Model} &
  Perception &
  \begin{tabular}[c]{@{}c@{}}Perception-\\ to-Belief\end{tabular} &
  \quad\ ToM\quad \  &
  Perception &
  \begin{tabular}[c]{@{}c@{}}Perception-\\ to-Belief\end{tabular} &
  \quad\ ToM\quad \  \\ \midrule
 &
  GPT-3.5 Turbo &
  \cellcolor[HTML]{FFFFFF}0.228 &
  \cellcolor[HTML]{C1D6F8}0.824 &
  \cellcolor[HTML]{BAD1F7}0.792 &
  \cellcolor[HTML]{FFFFFF}0.585 &
  \cellcolor[HTML]{D8E5FB}0.432 &
  \cellcolor[HTML]{F3F7FE}0.237 \\
 &
  GPT-4 Turbo &
  \cellcolor[HTML]{A4C2F4}0.934 &
  \cellcolor[HTML]{A4C2F4}0.980 &
  \cellcolor[HTML]{C5D8F8}0.739 &
  \cellcolor[HTML]{A4C2F4}0.950 &
  \cellcolor[HTML]{B5CEF7}0.723 &
  \cellcolor[HTML]{B3CCF6}0.780 \\
 &
  GPT-4o &
  \cellcolor[HTML]{A8C5F5}0.903 &
  \cellcolor[HTML]{BBD2F7}0.854 &
  \cellcolor[HTML]{D9E5FB}0.642 &
  \cellcolor[HTML]{ABC7F5}0.925 &
  \cellcolor[HTML]{A4C2F4}0.863 &
  \cellcolor[HTML]{A4C2F4}0.904 \\
 &
  Claude 3 Haiku &
  \cellcolor[HTML]{ACC8F5}0.874 &
  \cellcolor[HTML]{FFFFFF}0.480 &
  \cellcolor[HTML]{C6D9F9}0.730 &
  \cellcolor[HTML]{CADCF9}0.798 &
  \cellcolor[HTML]{B5CEF7}0.724 &
  \cellcolor[HTML]{ECF3FD}0.290 \\
 &
  Claude 3 Sonnet &
  \cellcolor[HTML]{ABC7F5}0.886 &
  \cellcolor[HTML]{A6C4F5}0.970 &
  \cellcolor[HTML]{A4C2F4}0.894 &
  \cellcolor[HTML]{B4CDF6}0.886 &
  \cellcolor[HTML]{DEE9FB}0.384 &
  \cellcolor[HTML]{EEF4FD}0.277 \\
 &
  Gemini 1.0 Pro &
  \cellcolor[HTML]{E6EEFC}0.425 &
  \cellcolor[HTML]{BCD2F7}0.850 &
  \cellcolor[HTML]{CFDFFA}0.690 &
  \cellcolor[HTML]{DBE7FB}0.733 &
  \cellcolor[HTML]{FFFFFF}0.104 &
  \cellcolor[HTML]{FFFFFF}0.127 \\
 &
  Llama-3 70B Instruct &
  \cellcolor[HTML]{B4CDF6}0.814 &
  \cellcolor[HTML]{C3D7F8}0.810 &
  \cellcolor[HTML]{FFFFFF}0.454 &
  \cellcolor[HTML]{DEE9FB}0.718 &
  \cellcolor[HTML]{E6EEFC}0.320 &
  \cellcolor[HTML]{B0CAF6}0.803 \\
\multirow{-8}{*}{\begin{tabular}[c]{@{}l@{}}Percept-\\ ToMi\end{tabular}} &
  Mixtral 8x22B Instruct &
  \cellcolor[HTML]{A6C4F5}0.920 &
  \cellcolor[HTML]{B4CDF6}0.894 &
  \cellcolor[HTML]{C4D7F8}0.743 &
  \cellcolor[HTML]{ADC8F5}0.917 &
  \cellcolor[HTML]{C3D7F8}0.607 &
  \cellcolor[HTML]{C8DBF9}0.597 \\ \midrule
 &
  GPT-3.5 Turbo &
  \cellcolor[HTML]{DCE8FB}0.866 &
  \cellcolor[HTML]{A4C2F4}0.505 &
  \cellcolor[HTML]{AEC9F6}0.177 &
  \cellcolor[HTML]{DAE7FB}0.877 &
  \cellcolor[HTML]{FFFFFF}0.000 &
  \cellcolor[HTML]{FFFFFF}0.000 \\
 &
  GPT-4 Turbo &
  \cellcolor[HTML]{AEC9F6}0.962 &
  \cellcolor[HTML]{E7EFFC}0.138 &
  \cellcolor[HTML]{D3E2FA}0.096 &
  \cellcolor[HTML]{AAC6F5}0.970 &
  \cellcolor[HTML]{C7DAF9}0.028 &
  \cellcolor[HTML]{A4C2F4}0.017 \\
 &
  GPT-4o &
  \cellcolor[HTML]{AAC6F5}0.970 &
  \cellcolor[HTML]{FCFDFF}0.020 &
  \cellcolor[HTML]{DCE8FB}0.077 &
  \cellcolor[HTML]{A6C4F5}0.977 &
  \cellcolor[HTML]{F3F7FE}0.006 &
  \cellcolor[HTML]{A4C2F4}0.017 \\
 &
  Claude 3 Haiku &
  \cellcolor[HTML]{FFFFFF}0.792 &
  \cellcolor[HTML]{FDFEFF}0.015 &
  \cellcolor[HTML]{F4F8FE}0.025 &
  \cellcolor[HTML]{FFFFFF}0.806 &
  \cellcolor[HTML]{EDF3FD}0.009 &
  \cellcolor[HTML]{F5F8FE}0.002 \\
 &
  Claude 3 Sonnet &
  \cellcolor[HTML]{A8C5F5}0.974 &
  \cellcolor[HTML]{FEFEFF}0.010 &
  \cellcolor[HTML]{FBFCFF}0.010 &
  \cellcolor[HTML]{A6C4F5}0.977 &
  \cellcolor[HTML]{EDF3FD}0.009 &
  \cellcolor[HTML]{FFFFFF}0.000 \\
 &
  Gemini 1.0 Pro &
  \cellcolor[HTML]{BAD1F7}0.937 &
  \cellcolor[HTML]{FFFFFF}0.000 &
  \cellcolor[HTML]{FFFFFF}0.000 &
  \cellcolor[HTML]{B4CDF6}0.950 &
  \cellcolor[HTML]{FBFDFF}0.002 &
  \cellcolor[HTML]{FFFFFF}0.000 \\
 &
  Llama-3 70B Instruct &
  \cellcolor[HTML]{A4C2F4}0.982 &
  \cellcolor[HTML]{EFF4FD}0.092 &
  \cellcolor[HTML]{A4C2F4}0.197 &
  \cellcolor[HTML]{A4C2F4}0.980 &
  \cellcolor[HTML]{D7E4FB}0.020 &
  \cellcolor[HTML]{DFEAFC}0.006 \\
\multirow{-8}{*}{\begin{tabular}[c]{@{}l@{}}Percept-\\ FANToM\end{tabular}} &
  Mixtral 8x22B Instruct &
  \cellcolor[HTML]{CCDDF9}0.899 &
  \cellcolor[HTML]{FEFEFF}0.010 &
  \cellcolor[HTML]{E8F0FD}0.051 &
  \cellcolor[HTML]{D3E1FA}0.892 &
  \cellcolor[HTML]{A4C2F4}0.045 &
  \cellcolor[HTML]{AFCAF6}0.015 \\ \bottomrule
\end{tabular}%
}}
\caption{LLM performances for perception inference, perception-to-belief inference, and Theory of Mind (ToM), as illustrated in Figure~\ref{fig:catplot_cap} for Percept-ToMi and Percept-FANToM.}
\label{tab:all_results}
\end{center}
\end{table*}

Lastly, based on the extracted perspective contexts, we build prompts to answer the ToM question.

\begin{mdframed}
\scriptsize{\texttt{Here are the past scenes in sequence that Lucas knows about. \\ \\
Lucas entered the cellar. The boots is in the cupboard. The cupboard is in the cellar. Lucas exited the cellar. Lucas entered the porch. \\ \\
Question: Where will Lucas look for the boots? State the most detailed position possible (e.g., in A in B). Answer in one sentence without explanation. \\
Answer: 
}}
\end{mdframed}

\begin{mdframed}
\scriptsize{\texttt{Here are the past utterances in sequence that Gianna is aware of. \\ \\
Gianna: Guys, I've really enjoyed sharing our pet stories, but I need to excuse myself. I need to change clothes for a meeting later. Talk to you later!\\
Sara: Sure thing, Gianna. Take care!\\
Javier: Catch you later, Gianna.\\
Gianna: Hey guys, I'm back, couldn't miss out on more pet stories. Speaking of teaching and training pets, it is amazing how that further strengthens the bond between us and our pets, right?\\
...\\ \\
Question: What does Gianna believe about who discussed their experiences training their pets, Bruno and Snowflake? Choose between (a) and (b). Do not include any explanation. \\
(a) Gianna believes that Sara and Javier discussed their experiences training their pets, Bruno and Snowflake. \\
(b) Gianna knows that Javier discussed training his pet, Bruno. However, Gianna will not know training a pet named Snowflake.
}}
\end{mdframed}

\section{LLM Performances on Percept-ToMi and Percept-FANToM}\label{appendix:catplot_cap_score}
Table~\ref{tab:all_results} presents the exact performance of Percept-ToMi and Percept-FANToM in perception inference, perception-to-belief inference, and ToM, which is also depicted in Figure~\ref{fig:catplot_cap}.

\section{Performance Comparison Between PercepToM and SymbolicToM}\label{sec:perceptom-symbolictom-comparison}

\begin{table*}[]
\begin{center}
\begin{tabular}{@{}cccc@{}}
\toprule
Model &
  Method &
  True Belief &
  False Belief \\ \midrule
\multirow{3}{*}{\begin{tabular}[c]{@{}c@{}}GPT-4 \\ Turbo\end{tabular}}       & PercepToM        & 0.824          & 1.000          \\
                                                                              & PercepToM+Oracle & 0.885          & \textbf{0.993} \\
                                                                              & SymbolicToM      & \textbf{0.997} & 0.977          \\ \midrule
\multirow{3}{*}{GPT-4o}                                                       & PercepToM        & 0.659          & 0.915          \\
                                                                              & PercepToM+Oracle & 0.660          & \textbf{0.993} \\
                                                                              & SymbolicToM      & \textbf{1.000} & 0.977          \\ \midrule
\multirow{3}{*}{\begin{tabular}[c]{@{}c@{}}Claude 3 \\ Sonnet\end{tabular}}   & PercepToM        & 0.963          & 0.937          \\
                                                                              & PercepToM+Oracle & 0.987          & \textbf{0.987} \\
                                                                              & SymbolicToM      & \textbf{1.000} & 0.977          \\ \midrule
\multirow{3}{*}{\begin{tabular}[c]{@{}c@{}}Llama-3 \\ 70B Inst.\end{tabular}} & PercepToM        & 0.713          & 0.744          \\
                                                                              & PercepToM+Oracle & 0.677          & \textbf{0.980} \\
                                                                              & SymbolicToM      & \textbf{1.000} & 0.977          \\ \midrule
\multirow{3}{*}{\begin{tabular}[c]{@{}c@{}}Mixtral \\ 8x22B Inst.\end{tabular}} &
  PercepToM &
  0.727 &
  0.964 \\
                                                                              & PercepToM+Oracle & 0.757          & 0.970          \\
                                                                              & SymbolicToM      & \textbf{1.000} & \textbf{0.977} \\ \bottomrule
\end{tabular}%
\caption{Performance comparison of PercepToM, PercepToM+Oracle, and SymbolicToM on the ToMi dataset. PercepToM+Oracle and PercepToM show comparable performance to SymbolicToM in false belief scenarios across most models. In true belief scenarios, SymbolicToM consistently outperforms PercepToM+Oracle, likely due to its question rephrasing process.}
\label{tab:perceptom-symbolictom}
\end{center}
\end{table*}

Table~\ref{tab:perceptom-symbolictom} shows the performances of PercepToM, PercepToM+Oracle, and SymbolicToM on ToMi. 

\section{Performance Comparison Between PercepToM and SimToM}\label{sec:perceptom-simtom-comparison}

\begin{table*}[]
\begin{center}
\begin{tabular}{@{}cccc@{}}
\toprule
Model                & Method    & True Belief    & False Belief   \\ \midrule
GPT-4  Turbo         & SimToM    & 0.657          & 0.873          \\
                     & PercepToM & \textbf{0.824} & \textbf{1.000} \\ \midrule
GPT-4o               & SimToM    & \textbf{0.797} & 0.450          \\
                     & PercepToM & 0.659          & \textbf{0.915} \\ \midrule
Llama-3 70B Inst.    & SimToM    & 0.644          & \textbf{0.770} \\
                     & PercepToM & \textbf{0.713} & 0.744          \\ \midrule
Mixtral  8x22B Inst. & SimToM    & 0.677          & 0.660          \\
                     & PercepToM & \textbf{0.727} & \textbf{0.964} \\ \bottomrule
\end{tabular}
\caption{Performance comparison between SimToM and PercepToM on Fixed and Disambiguated ToMi~\citep{sclar-etal-2023-minding}. Overall, PercepToM shows more robust performance across different models in both of true and false belief scenarios.}
\label{tab:perceptom_simtom}
\end{center}
\end{table*}

Table~\ref{tab:perceptom_simtom} shows the performances of PercepToM and SimToM on ToMi.

\label{sec:appendix}

\end{document}